\documentclass[runningheads]{llncs}
\usepackage[T1]{fontenc}
\usepackage{graphicx}
\usepackage{url}

\begin{document}
%
\title{GazeSAM: What You See is What You Segment}

\author{Bin Wang, Armstrong Aboah, Zheyuan Zhang, Ulas Bagci}
\authorrunning{Wang et al.}
%
\institute{Northwestern University, Chicago IL, 60611, USA \\
\email{ulas.bagci@northwestern.edu}}
\maketitle              
\begin{abstract}
This study investigates the potential of eye-tracking technology and the Segment Anything Model (SAM) to design a collaborative human-computer interaction system that automates medical image segmentation. We present the \textbf{GazeSAM} system to enable radiologists to collect segmentation masks by simply looking at the region of interest during image diagnosis. The proposed system tracks radiologists' eye movement and utilizes the eye-gaze data as the input prompt for SAM, which automatically generates the segmentation mask in real time. This study is the first work to leverage the power of eye-tracking technology and SAM to enhance the efficiency of daily clinical practice. Moreover, eye-gaze data coupled with image and corresponding segmentation labels can be easily recorded for further advanced eye-tracking research. The code is available in \url{https://github.com/ukaukaaaa/GazeSAM}.

\keywords{Eye-Tracking Technology  \and Segment Anything \and Human-Computer Interaction.}
\end{abstract}
%
%

\section{Introduction}

Image segmentation is a crucial process in numerous medical applications, such as disease diagnosis, treatment planning, and surgical navigation. It involves the identification of regions of interest (ROIs) in medical images, such as organs, tumors, and lesions \cite{altini2022liver,florez2018emergence,tunali2021application}. Accurate segmentation helps radiologists to understand a patient's condition better and to develop more effective treatment plans. However, the segmentation of medical images has primarily been accomplished through a manual annotation process. This is a costly and time-consuming procedure that can take hours or days to complete. This bottleneck presents a significant barrier to the widespread adoption of image segmentation in clinical practice.

As a result, there is a growing need for more intelligent and efficient approaches to segmenting medical images. One promising approach which has not yet been explored is the use of eye-tracking technology to perform image segmentation in real time. While previous research in eye tracking has primarily focused on understanding the relationship between human attention and cognitive decision-making \cite{wood2020eye,khosravan2019collaborative}, its potential in automating the segmentation of medical images has yet to be fully realized. By tracking radiologists' eye movements when they read medical images, it is possible to identify the ROIs that are most relevant to them using their gaze points. This rich information can be leveraged to segment the images automatically.

While the Segment Anything Model (SAM) has shown tremendous success across various domain applications, its potential to form part of a framework that creates an interactive system for radiologists to segment medical images has not been probed in by earlier studies. As such, there is a need to investigate the possibility of developing a collaborative and interactive framework that can assist radiologists in automatically annotating medical images in real time.

Hence, we propose the GazeSAM system to investigate the feasibility and efficacy of integrating eye-tracking technology and the SAM into a collaborative AI-assisted system for real-time medical image segmentation. The system uses eye-tracking technology to identify the ROIs that radiologists are interested in and then prompts the SAM model to segment the images accordingly with the eye-gaze points. The system is designed to be user-friendly, accurate, and fast in generating segmentation results. It is worth noting that this is the first study to utilize the power of eye-tracking data and SAM to automate the segmentation process of medical images in real time. 

The major contributions of this work are summarized as follows:
\begin{enumerate}
    \item We propose a collaborative human-computer interaction system \textbf{GazeSAM} that combines eye-tracking technology with SAM for real-time medical image segmentation by radiologists.
    \item GazeSAM system employs a screen-based eye tracker that offers superior accuracy and greater comfort to radiologists compared to the glass-based eye trackers employed by the Meta Virtual Reality Team. Moreover, most eye-tracking datasets are collected using screen-based eye trackers, making our system more suitable for standard eye-tracking dataset collection.
    \item Our system has the unique capability of operating with both 2D and 3D images, which are typically used in medical settings. This is the first of its kind developed to increase radiologists' work efficiency in daily clinical practice significantly.
    \item The system can be employed for both coarse segmentation mask collection and the accompanying acquisition of eye-tracking data, which can foster advancements in further eye-tracking research.
\end{enumerate}

Despite the fact that previous studies have indicated that the Segment Anything Model (SAM) may not be effective in segmenting medical images \cite{roy2023sam,deng2023segment}, the current research is not primarily focused on evaluating the model's performance. Instead, the research objective is to investigate the potential of SAM as a system within a framework for real-time medical image segmentation in a collaborative and interactive context. The results from this work have the potential to open up new research areas and improve the real-time application of SAM in medical image segmentation in collaborative environments, ultimately contributing to the development of more efficient and accurate methods for medical image analysis and potentially enhancing clinical practice and research.


\section{Method}
\begin{figure}[htbp]
\includegraphics[width=\textwidth]{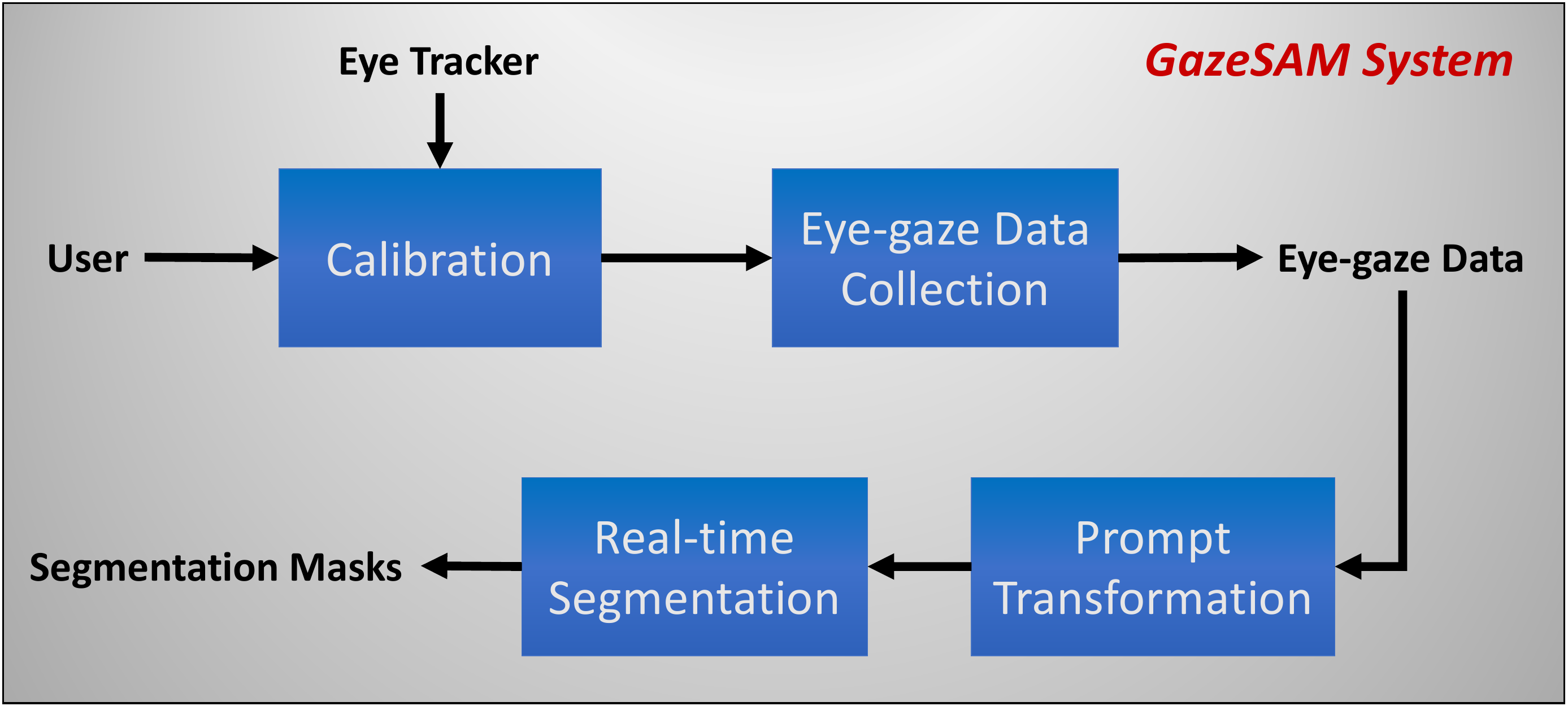}
\caption{Overview of our proposed system.} \label{framework}
\end{figure}

In this section, we describe our proposed framework GazeSAM for real-time segmentation mask collection by utilizing a screen-based eye-tracker and Segment-Anything Model (SAM). As illustrated in Fig.~\ref{framework}, GazeSAM comprises two parts: eye-gaze data collection and segmentation. 

\subsection{Eye-gaze Data Collection}
In this study, a Tobii Pro Nano screen-based eye tracker is used. It is a small, lightweight, and easy-to-use eye tracker whose length is 170mm, weight is 59g, and the sampling rate is 60Hz.

Before the experiment, calibration of the eye tracker is required because it ensures the eye movement is tracked accurately and makes the gaze coordinate on the screen consistent with where the user is looking. Here, we adopt a five-point calibration procedure in Tobii Pro eye tracker manager. After completing the calibration, the eye-gaze data can be collected in the form of the location coordinate on the screen.

\subsection{Prompt Transformation \& Segmentation} \label{optionsec}
The prompt encoder in SAM is designed to support a wide range of prompt formats, such as points, boxes, and text. To integrate eye-gaze data as a new type of prompt into the SAM, we need to first conduct a prompt transformation.

\begin{figure}[htbp]
\includegraphics[width=\textwidth]{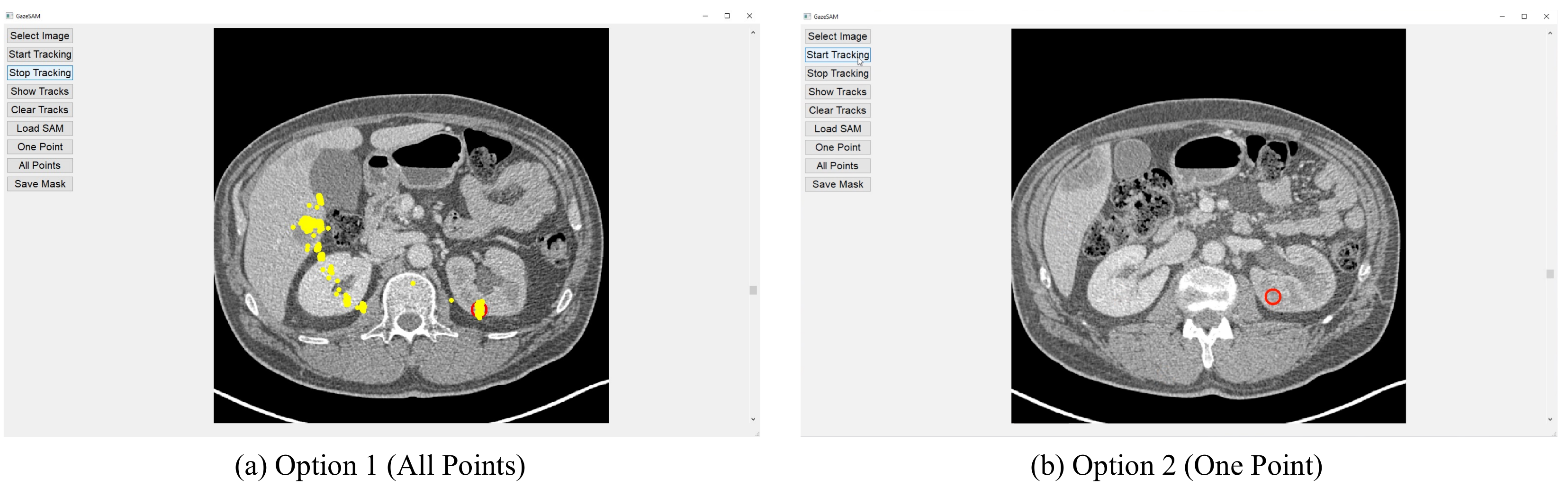}
\caption{Two eye-gaze prompt options for segmentation in GazeSAM.} \label{option}
\end{figure}

Eye-gaze data can be considered as a sequence of scatter points that correspond to the eye movement over time. Hence, it is possible to transfer the eye-gaze data into a point or a sequence of points, which can be utilized as the point prompt for SAM. Prior to this, it is necessary to first solve the coordinate problem. The eye-gaze points coordinates, denoted as $S_1, S_2,... S_n$  are collected in the screen coordinate space. We need to transform it into the image coordinate space as follows:
\begin{equation}
    I_1, I_2, ..., I_n = f(S_1, S_2,... S_n),
\end{equation}
where $f(\cdot)$ is the mapping function between two coordinate space and $I_1, I_2, ..., I_n$ are the eye-gaze points coordinates in image coordinate space.

Then, as illustrated in Fig.~\ref{option}, GazeSAM supports two options for inputting the eye-gaze data as a prompt for SAM. The first option is to use the whole sequence of eye-gaze points collected over time, which can provide a more comprehensive representation of the user's gaze trajectory. The second option is to use the eye-gaze point collected at the last time point as the prompt. This approach is more appropriate when a coarse segmentation mask of a single object is desired.

It is noted that SAM might not always generate a perfect segmentation mask, especially for the boundary regions. To refine generated mask, users need to manually add points to those regions, which can be tedious and time-consuming. In the first option, GazeSAM simplifies this process by allowing users to add points by simply looking at the desired areas. In this way, a more efficient approach to refine the segmentation mask is offered, which has the potential to greatly enhance the user experience and speed of the whole pipeline.

Given a pre-computed image embedding and the prompt transformed from eye-gaze data, SAM can generate a segmentation mask subsequently in near real-time, making it an interactive segmentation system by using eye-tracking technology.


\section{Experiments}
\begin{figure}
\includegraphics[width=\textwidth]{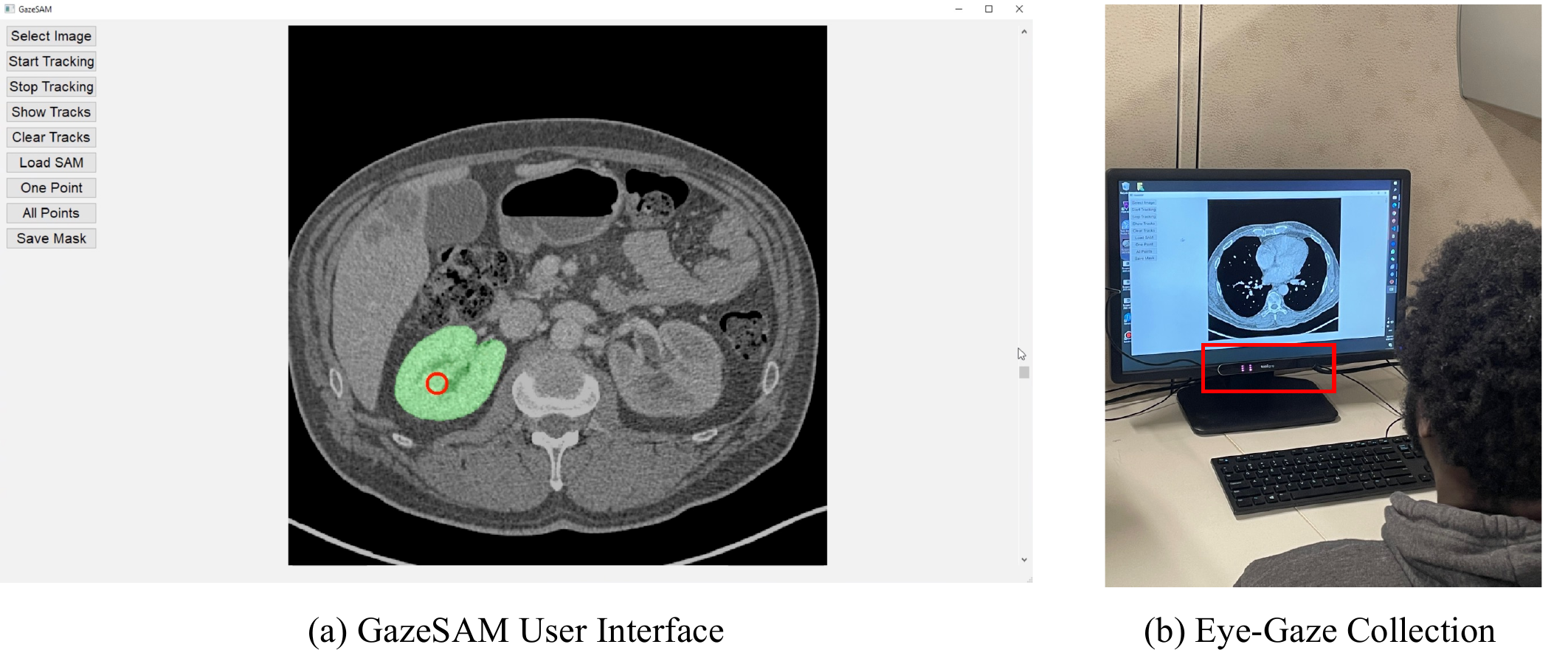}
\caption{Our designed user interface and experiment setting of eye-gaze data collection.} \label{setting}
\end{figure}
As illustrated in Fig.~\ref{setting}(b), the eye tracker is positioned directly below the lower edge of the display, and the user maintains a viewing distance of approximately 60cm from the screen.

To visualize the eye-gaze points and the following segmentation result efficiently, we have developed a user interface that incorporates multiple functions as illustrated in Fig.~\ref{setting}(a). The function panel is situated on the left side of the interface, and each function can be easily activated either using corresponding keyboard shortcuts or clicking the buttons. The "Select Image" button enables users to choose the image they wish to segment, and it supports both 2D and 3D images. For 3D medical images, a scroll bar is provided on the right side of the interface to manipulate the slice image. 
\begin{figure}[h]
\includegraphics[width=\textwidth]{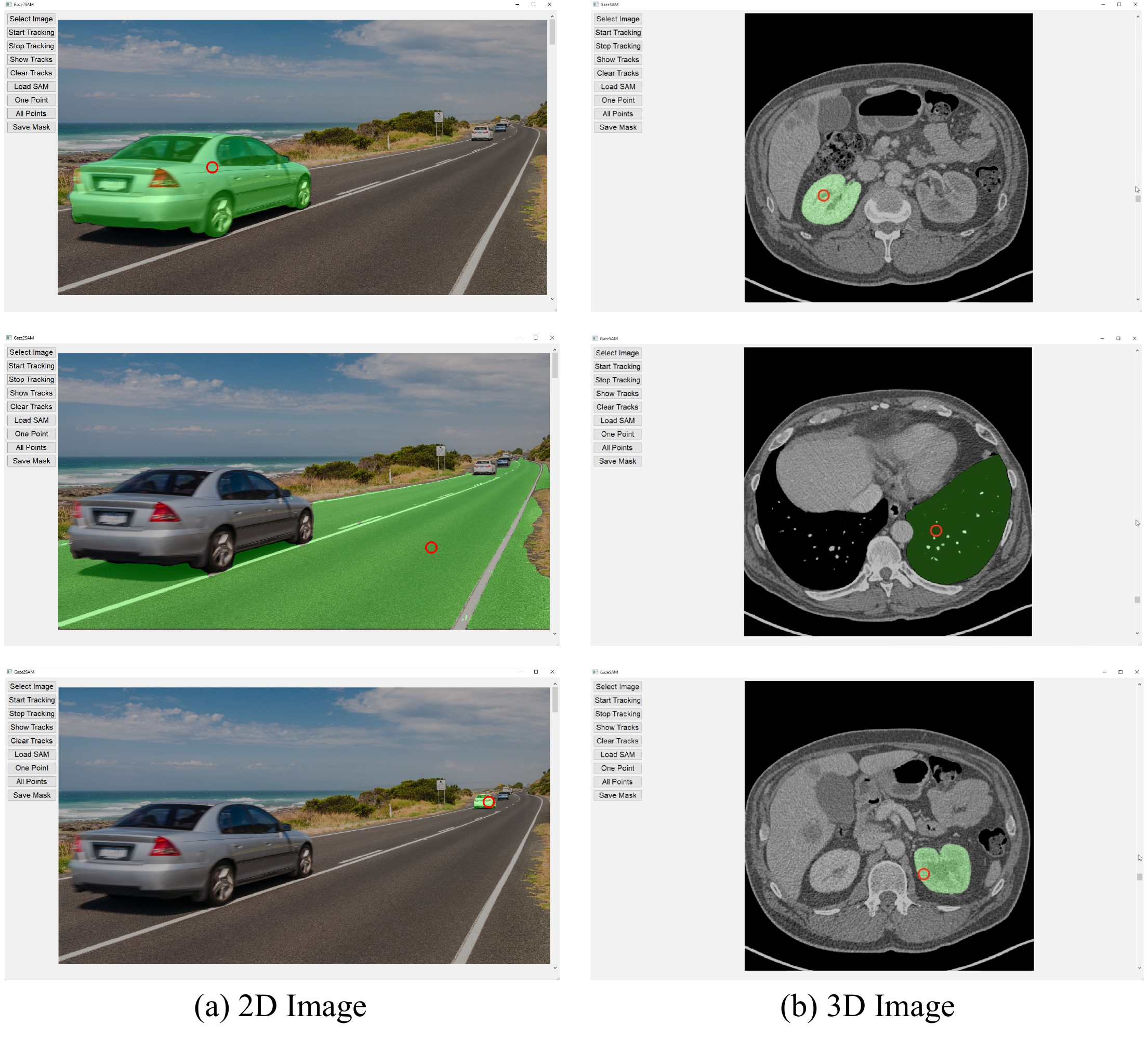}
\caption{Visualization of GazeSAM with 2D image and 3D image.} \label{2d3d}
\end{figure}

Once the user clicks the "Start Tracking" button, the eye-gaze point is displayed as a hollow red circle, which tracks the user's eye movement in real time. The red circle will follow the trajectory of the user's eye as it moves across the screen. The "Show Tracks" option displays the eye movement trajectory as yellow dots in Fig.~\ref{option}(a). 

For segmentation, GazeSAM provides two options ("One Point" and "All Points") as described in Section~\ref{optionsec}. Once an option is activated, a segmentation mask is generated in real-time and shown in green color. This mask is automatically updated based on the location of the user's eye gaze, allowing for dynamic adjustments to the segment target or iterative refinement of the segmentation. Users can save the eye-captured segmentation mask displayed on the interface by using the "Save Mask" function at any point during the process.



\section{Discussion}
Given that SAM is primarily trained on natural images, its ability to infer accurate segmentation on medical images is limited. While GazeSAM offers a more efficient approach to improve segmentation quality by simply looking at the desired areas and incorporating more eye-gaze prompts in regions with poor segmentation, its performance is still restricted in some cases. To overcome this limitation, fine-tuning SAM on a large-scale medical image dataset is a possible solution~\cite{ma2023segment}. Besides, we can also use the generated mask as a coarse segmentation result for further refinement.

\newpage
\bibliographystyle{splncs04}
\bibliography{ref}
\end{document}